%% file: main.tex
\documentclass[10pt,twocolumn,letterpaper]{article}

\usepackage[pagenumbers]{iccv} 


\newcommand{\method}{FDINet\xspace}
\usepackage{subcaption}
\usepackage{amssymb}
\usepackage{amsmath}
\usepackage{multirow}

\usepackage{xspace}
\usepackage{enumitem}
\usepackage{graphicx}  
\usepackage{caption} 
\usepackage{amsmath}
\usepackage{wrapfig}
\usepackage{multirow}
\usepackage{colortbl}
\usepackage{pifont}
\usepackage{algorithm}
\usepackage{algorithmicx}
\usepackage[noend]{algpseudocode}
\definecolor{iccvblue}{rgb}{0.21,0.49,0.74}
\usepackage[pagebackref,breaklinks,colorlinks,allcolors=iccvblue]{hyperref}

\def\paperID{3414} 
\def\confName{ICCV}
\def\confYear{2025}

\title{Efficient Continual Learning through Frequency Decomposition and Integration}

\author{
  Ruiqi Liu\text{$^{1,2}$}, Boyu Diao\text{$^{1,2,\dagger}$}, Libo Huang\text{$^{1}$}, Hangda Liu\text{$^{1,2}$},\\
  Chuanguang Yang\text{$^{1,2}$}, Zhulin An\text{$^{1,2}$}, Yongjun Xu\text{$^{1,2}$} \\
  $^{1}$Institute of Computing Technology, Chinese Academy of Sciences, Beijing, China\\
  $^{2}$University of Chinese Academy of Sciences, Beijing, China\\
  \texttt{liuruiqi23@mails.ucas.ac.cn},\\
  \texttt{\{diaoboyu2012,	liuhangda21s, yangchuanguang, anzhulin, xyj\}@ict.ac.cn}, \\
  \texttt{www.huanglibo@gmail.com} \\
}

\begin{document}
\maketitle
\def\thefootnote{$\dagger$}\footnotetext{Corresponding Author.}
\begin{abstract}
Continual learning (CL) aims to learn new tasks while retaining past knowledge, addressing the challenge of forgetting during task adaptation. Rehearsal-based methods, which replay previous samples, effectively mitigate forgetting. However, research on enhancing the efficiency of these methods, especially in resource-constrained environments, remains limited, hindering their application in real-world systems with dynamic data streams. The human perceptual system processes visual scenes through complementary frequency channels: low-frequency signals capture holistic cues, while high-frequency components convey structural details vital for fine-grained discrimination. Inspired by this, we propose the Frequency Decomposition and Integration Network (\method), a novel framework that decomposes and integrates information across frequencies. \method designs two lightweight networks to independently process low- and high-frequency components of images. When integrated with rehearsal-based methods, this frequency-aware design effectively enhances cross-task generalization through low-frequency information, preserves class-specific details using high-frequency information, and facilitates efficient training due to its lightweight architecture. Experiments demonstrate that \method reduces backbone parameters by 78\%, improves accuracy by up to 7.49\% over state-of-the-art (SOTA) methods, and decreases peak memory usage by up to 80\%. Additionally, on edge devices, \method accelerates training by up to 5$\times$.
\end{abstract}
\section{Introduction}
\label{sec:intro}
Continual learning (CL) allows machine learning models to adapt to new data while retaining prior knowledge in dynamic environments~\cite{kirkpatrick2017overcoming}. Nonetheless, the absence of previous training data can result in \textit{catastrophic forgetting}~\cite{mccloskey1989catastrophic}, since fitting new classes can erase patterns from previous ones, resulting in performance decline. While recent CL methods focus on preventing forgetting, it is crucial to prioritize learning efficiency on resource-constrained edge devices~\cite{pellegrini2021continual}, such as smartphones, smart cameras, and embedded systems like the NVIDIA Jetson Orin NX. This becomes especially critical in applications constrained by privacy requirements, limited network connectivity, and the need for rapid adaptation~\cite{pellegrini2020latent,liu2024resource}.
\begin{figure} [t!]
     \centering
     \includegraphics[width=0.9\linewidth]{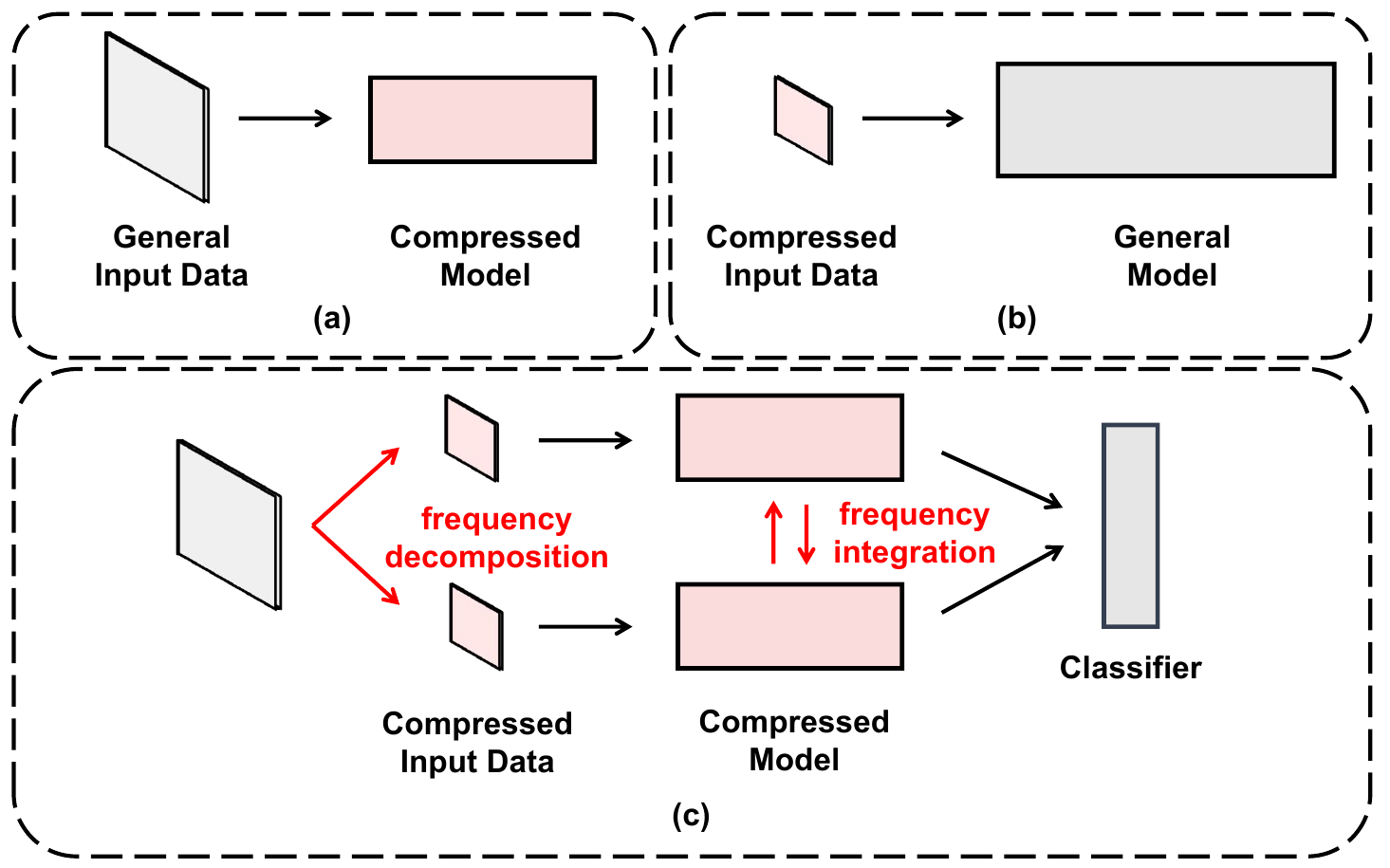}  
     \caption{(a) Diagram of the model compression method. (b) Diagram of the input compression method. (c) Diagram of the proposed \method. By utilizing frequency decomposition and integration, \method compresses both the input and the model. Through the synergistic replay mechanism of high- and low-frequency components, it compensates for the potential performance loss caused by compression.}
    \label{fig:over}  
\end{figure}
\begin{figure*} [t]
     \centering
     \includegraphics[width=0.7\linewidth]{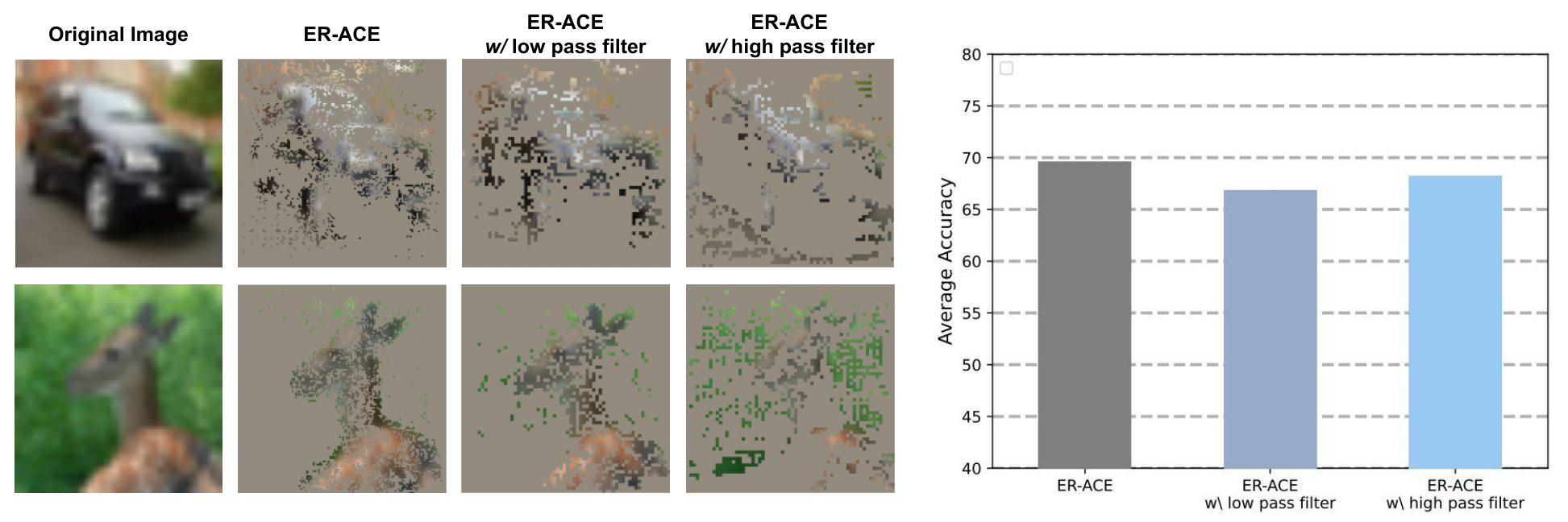}  
     \caption{Continual learning results on Split CIFAR-10 (5 tasks) using ER-ACE~\cite{caccia2021new}. Images were pre-processed with high-pass and low-pass filters. \textbf{Left:} Gradient-based attention maps indicating model-attended input pixels. Low-frequency features in the images tend to focus on global structures, while high-frequency features focus more on edge textures and are susceptible to noise. \textbf{Right:} Classification accuracy. Despite filtering out a significant amount of image information, the accuracy degradation is not substantial.}
    \label{fig:high_low}  
\end{figure*}
To mitigate \textit{catastrophic forgetting}, various methods have been employed: \textit{regularization-based} methods~\cite{huang2024etag,serra2018overcoming,li2017learning,chaudhry2018riemannian,huang2024kfc} limit changes to key parameters, preserving crucial knowledge from previous tasks; \textit{architecture-based} methods~\cite{ebrahimi2020adversarial,ke2020continual,loo2020generalized,wang2020learn,zhao2022deep} assign distinct parameters to each task or expand the network to prevent interference between tasks; and \textit{rehearsal-based} methods~\cite{aljundi2019gradient,buzzega2020dark,cha2021co2l,chaudhry2021using,chaudhry2018efficient} replay stored samples to maintain past knowledge. Rehearsal-based methods have proven to be the most effective~\cite{buzzega2020dark}. However, training efficiency and memory footprint are critical concerns for deploying these methods on edge devices. Long training times and limited edge memory render existing methods impractical for real-world edge applications~\cite{prabhu2023computationally,ghunaim2023real}.  Therefore, prioritizing inherent method efficiency over complex optimizations is essential for efficient CL method design. To address this, some studies have explored training efficiency in rehearsal-based methods. SOTA efficient CL methods can be broadly classified into two categories: model compression~\cite{wang2022sparcl} and input compression~\cite{liu2024continual}. As illustrated in Fig.~\ref{fig:over} (a), model compression~\cite{wang2022sparcl} reduces the number of network parameters through sparse training. However, this method fails to effectively reduce training memory footprint because the main bottleneck is intermediate activation storage, not parameters~\cite{chen2016training,cai2020tinytl,yang2022da3}. Input compression~\cite{liu2024continual}, as shown in Fig.~\ref{fig:over} (b), reduces memory footprint and speeds up training by decreasing the size of input feature maps. However, this method fails to fully exploit the redundancy in network parameters, leading to extra computational and memory overhead. To the best of our knowledge, no existing method effectively compresses both input and model concurrently. This motivates our exploration of a joint compression method to improve CL efficiency.

Research on the human perception system indicates that both low- and high-frequency components are critical for visual processing. Low-frequency components capture global structures for robust classification, while high-frequency components encode local details for fine-grained recognition~\cite{walther2011simple,huang2017wavelet}. Fig.~\ref{fig:high_low} shows the impact of image frequency components on CL. Analogous to the human perception system, high-frequency components in CL capture class-specific edges but are sensitive to noise. In contrast, low-frequency components capture global structure, sacrificing some class details and potentially causing \textit{catastrophic forgetting}. Notably, despite substantial information filtering, accuracy loss remains modest in both low- and high-pass filtering.

This observation prompts the question: \textbf{Could decoupling high- and low-frequency feature memorization accelerate training with smaller inputs and lighter models without sacrificing accuracy?} This method may leverage low-frequency components for general class features and high-frequency components to reinforce specific details during replay.

To this end, we propose the Frequency Decomposition and Integration Network (\method), a novel CL framework that simultaneously optimizes the training efficiency and performance of CL methods. As shown in Fig.~\ref{fig:over} (c), \method accelerates training and improves accuracy through the decomposition and integration of different frequency components of the image. Specifically, we design two filters to decompose the input feature maps into reduced-size high-frequency and low-frequency input features.  These smaller input features are then processed by two lightweight networks to extract their respective features, which are subsequently classified. To promote more efficient utilization of different frequency features, we design feature aggregators to integrate the intermediate high- and low-frequency features from the two lightweight networks. Our rehearsal framework enhances holistic representations via low-frequency preservation and mitigates the forgetting of class-specific information through high-frequency retention, while significantly compressing the model’s input and structure. Therefore, when combined with various rehearsal-based methods, \method can significantly reduce their memory footprint, accelerate training, and maintain accuracy, greatly promoting CL on edge devices.
In summary, our contributions are as follows:
\begin{itemize}
\item We propose \method, a novel CL framework designed to collaboratively optimize the training efficiency and performance of CL methods. \method compresses the network input and structure through the decomposition and integration of image frequencies.
\item We propose a frequency-decoupled rehearsal strategy that preserves class stability through low-frequency prototypes while enhancing discriminative power via high-frequency details, thereby mitigating compression-induced performance degradation.
\item \method outperforms SOTA methods in extensive experiments. It boosts accuracy by up to 7.49\%, slashes backbone parameters by 78\%, reduces peak memory by 80\%, and accelerates training by 5$\times$ on edge devices.
\end{itemize}

\section{Related Work}
\label{sec:related}

\paragraph{Effective Continual Learning.}
Effective continual learning focuses on addressing the problem of \textit{catastrophic forgetting}, typically divided into three main categories: regularization-based, architecture-based, and rehearsal-based methods.

\textit{Regularization-based Methods}~\cite{yu2020semantic,serra2018overcoming,li2017learning,yang2024clip,chaudhry2018riemannian,aljundi2018memory} constrain the changes to critical parameters learned from previous tasks to mitigate forgetting. However, their reliance on soft penalties is insufficient to fully prevent forgetting, especially in complex datasets or when dealing with a large number of tasks, resulting in poor performance in class-incremental learning scenarios~\cite{buzzega2020dark}.

\textit{Architecture-based Methods}~\cite{ebrahimi2020adversarial,ke2020continual,loo2020generalized,wang2020learn,zhao2022deep} allocate separate parameter sets to each task, either by expanding the network or dividing existing components. However, these methods often suffer from capacity saturation and scalability issues as the number of tasks increases. Continual expansion of modules can lead to significant computational and memory overhead, making them impractical for long-term incremental learning~\cite{zhu2021class,wang2022foster}.

\begin{figure*} [t]
     \centering
     \includegraphics[width=0.8\linewidth]{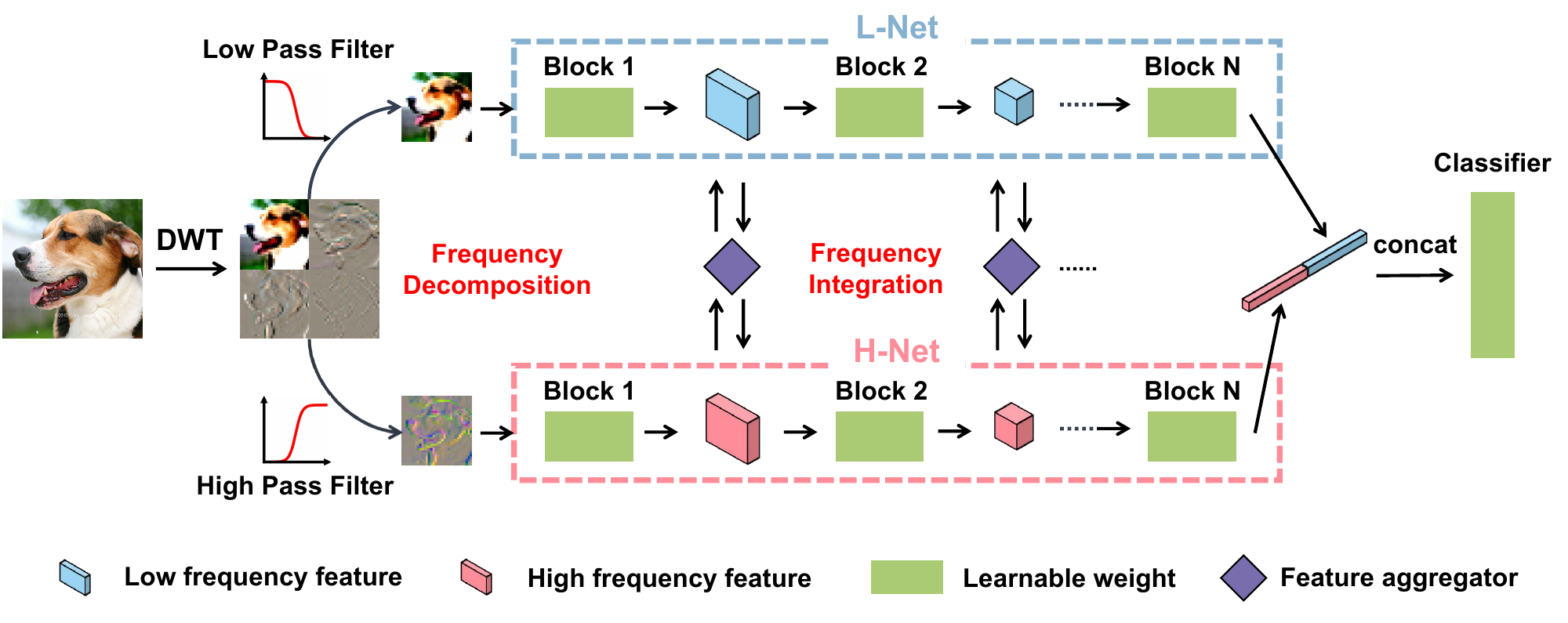}  
     \caption{The main framework of \method.
        We decompose the original image into frequency components using discrete wavelet transform. Subsequently, two lightweight networks are used to extract features from the reduced-size high-frequency and low-frequency inputs, respectively. To better leverage the information from different frequencies, we design feature aggregators to integrate the intermediate frequency features from both lightweight networks.}
    \label{fig:method}  
\end{figure*}

\textit{Rehearsal-based Methods}~\cite{aljundi2019gradient,buzzega2020dark,cha2021co2l,chaudhry2021using,chaudhry2018efficient} mitigate forgetting by maintaining an episodic memory buffer that stores a small subset of previous tasks' data, which are replayed alongside new task data to approximate the joint distribution during training. Our framework focuses on rehearsal-based methods, as they are widely regarded as the most effective for mitigating \textit{catastrophic forgetting}~\cite{buzzega2020dark}.
ER~\cite{robins1995catastrophic} is a simple yet effective rehearsal method that stores past examples in a fixed memory buffer and replays them during training. Building on this concept, DER++~\cite{buzzega2020dark} retains past model logits as soft targets, applying a consistency loss during training to improve learning. ER-ACE~\cite{caccia2021new} reduces representation drift after task switches by using asymmetric update rules. Moreover, CLS-ER~\cite{arani2022learning} mimics complementary learning systems by maintaining two semantic memories that aggregate model weights at different timescales.
\paragraph{Efficient Continual Learning.}
Efficient continual learning~\cite{wang2022sparcl,lee2018snip,evci2020rigging,liu2024continual} aims to improve training speed, reduce memory footprint, and optimize computational resource utilization while keeping the model size fixed and avoiding unnecessary computational overhead, making it suitable for real-time and resource-constrained scenarios. Some works~\cite{gao2024beyond,tang2024mind,kim2024eclipse} optimize transformer architectures for training efficiency but are hard to integrate with other CL methods, serving as independent CL methods rather than a unified framework for enhancing CL. Other works~\cite{harun2023siesta,hayes2020remind,pellegrini2020latent,pellegrini2021continual} accelerate training and reduce memory footprint by adjusting the replay position within the model, but they are prone to aging effects~\cite{pellegrini2020latent}. SparCL~\cite{wang2022sparcl} compresses the network structure using dynamic weight and gradient masks, combined with data filtering, thereby effectively reducing training FLOPs.
CLFD~\cite{liu2024continual} compresses the input by mapping images to the frequency domain and selectively utilizing frequency domain features, thereby optimizing training speed and memory footprint in CL. However, they lack explicit decomposition of high- and low-frequency components, instead processing full-spectrum signals through basic data-level transformations, which neglects the potential of leveraging distinct frequency components to mitigate \textit{catastrophic forgetting} and enhance overall performance. The primary distinction from previous work lies in our introduction of a frequency-decomposed replay mechanism. \method proposes the separate memorization of high- and low-frequency image features to achieve compelling CL performance, even with smaller input dimensions and lightweight models.
\section{Method}
\label{sec:method}
We describe a novel framework, \method (as shown in Fig.~\ref{fig:method}), in this section. \method optimizes the training efficiency of CL methods through the decomposition and integration of frequency components. Before going into in-depth, we first clarify the problem setting.

\subsection{Problem Setting}
\label{sec:method:overview}
CL involves sequentially learning from a sequence of \( T \) tasks, denoted as \( \{ \mathcal{D}^1, \ldots, \mathcal{D}^T \} \), where each task \( t \) consists of a dataset \( \mathcal{D}^t = \{(x_i, y_i)\}_{i=1}^{N_t} \). Each dataset contains pairs of input samples \( x_i \) and their corresponding labels \( y_i \), drawn from a task-specific data distribution. During CL, the model faces the challenge of learning new information while retaining knowledge from earlier tasks, which often leads to \textit{catastrophic forgetting}.
To address this, rehearsal-based methods maintain a memory buffer \( \mathcal{M} = \{(x_i, y_i)\}_{i=1}^{\mathcal{B}} \), which stores a small subset of previously observed data. The memory buffer is of limited size, \( \mathcal{B} \ll N_t \), reflecting the inherent constraints of CL in terms of memory resources. To efficiently manage this buffer, we employ reservoir sampling~\cite{vitter1985random}, which allows for a balanced representation of past tasks.

\subsection{Frequency Decomposition}
\label{sec:method:Decomposition}
In CL, low-frequency components capture shared features across classes, while high-frequency components convey class-specific information\cite{zhao2021mgsvf}. Replaying entire images may lead to interference between different frequencies. Therefore, to mitigate \textit{catastrophic forgetting}, we propose processing these components separately: blurring low frequencies to preserve shared structural information and sharpening high frequencies to enhance class-specific details. To achieve this separation, we perform frequency decomposition to compress the input. Initially, we utilize the Discrete Wavelet Transform (DWT) to map the input image to the frequency domain. Owing to its effective signal representation in both spatial and frequency domains, DWT is more suitable for CL compared to the Discrete Cosine Transform (DCT) and Discrete Fourier Transform (DFT)~\cite{liu2024continual}.
For a 2D signal $X \in \mathbb{R}^{N \times N}$, the DWT decomposes the signal into one low-frequency component, $X_{ll}$, and three high-frequency components, $X_{lh}$, $X_{hl}$, $X_{hh}$, represented as follows:
\begin{equation}
\begin{split}
    X_{ll} &= L X L^T, \\
    X_{lh} &= L X H^T, \\
    X_{hl} &= H X L^T, \\
    X_{hh} &= H X H^T,
\end{split}
\end{equation}
where $L$ and $H$ are the low-frequency and high-frequency filters of orthogonal wavelets, truncated to the size $\left\lfloor\frac{N}{2}\right\rfloor \times N$, which can be represented as follows:
\begin{equation}
\mathbf{L} = \begin{pmatrix}
\cdots & \cdots\\
\cdots & l_{0} & l_1 & l_2 & \cdots & & \\
& &\cdots & l_{0} & l_1 & l_2 & \cdots \\
& & & & &\cdots &\cdots\\
\end{pmatrix}, 
\end{equation}
\begin{equation}
\mathbf{H} = \begin{pmatrix}
\cdots & \cdots\\
\cdots & h_{0} & h_1 & h_2 & \cdots & & \\
& &\cdots & h_{0} & h_1 & h_2 & \cdots \\
& & & & &\cdots &\cdots\\
\end{pmatrix}.
\end{equation}
We choose the Haar wavelet as the basis for the DWT because of its computational efficiency~\cite{levinskis2013convolutional}. Specifically, the filters are given by $\mathbf{l} = \frac{1}{\sqrt{2}} \{1, 1\}$ and $\mathbf{h} = \frac{1}{\sqrt{2}} \{1, -1\}$. To obtain downsampled low-frequency and high-frequency input features, we use a low-pass filter (filtering $X_{lh}, X_{hl}, X_{hh}$ frequency components) and a high-pass filter (filtering the $X_{ll}$ frequency component and using a $1 \times 1$ point convolution to fuse all high-frequency components). These input features are then fed into the low-frequency feature extraction network (L-Net) and the high-frequency feature extraction network (H-Net), respectively. The extracted features are subsequently concatenated and input into the classifier for final classification. We only store the downsampled high-frequency and low-frequency input features, which allows us to store more features with the same buffer size.

Balancing network dimensions—including width, depth, and resolution—is crucial. In designing L-Net and H-Net, we halve both the depth (i.e., the number of convolutional layers) and the width (i.e., the number of convolutional channels) of the backbone network, which allows the use of lightweight networks for feature extraction. We also consider two additional scenarios: halving only the depth or only the width. An ablation study of these scenarios is presented in Tab.~\ref{tab:Ablation}. The results in rows 5 to 6 indicate that increasing the network width and depth does not effectively improve performance, underscoring the importance of efficiently balancing different network dimensions. It is also noteworthy that increasing depth leads to a performance drop when combined with CLS-ER, which is consistent with previous findings~\cite{mirzadeh2022wide,guha2024diminishing} that adding depth may increase the risk of \textit{catastrophic forgetting}. We analyze the computational complexity of \method in Appendix A.
\begin{table*}[t]
\centering
\renewcommand\arraystretch{1.4}\setlength{\tabcolsep}{1.1mm}
\caption{We evaluate the impact of different designs of feature aggregators and lightweight networks on \method's performance using the Split CIFAR-10 dataset with a buffer size of 125. Results of our framework are highlighted in bold.}
	\begin{tabular}{l|ccc|ccc}
\toprule
		\multirow{3}{*}{\small{{\bf  Method}}}&\multicolumn{3}{c|}{\small{{\bf Combined with ER}}}&\multicolumn{3}{c}{\small{{\bf Combined with CLS-ER}}}\\
		\cmidrule{2-7}
		&\multirow{2}{*}{\small{Class-IL($\uparrow$)}}&\multirow{2}{*}{\small{Task-IL($\uparrow$)}}& \small{FLOPs Train}&\multirow{2}{*}{\small{Class-IL($\uparrow$)}}&\multirow{2}{*}{\small{Task-IL($\uparrow$)}}& \small{FLOPs Train} \\
		&&&\small{$\times 10^{15}$($\downarrow$)}&&&\small{$\times 10^{15}$($\downarrow$)}\\
        \midrule
        No Integration (a)& 37.58\tiny±3.07 & 88.46\tiny±0.41 & 0.64& 56.02\tiny±0.81 & 90.18\tiny±0.28 & 0.96 \\
        Low-Frequency Dominance (b) & 44.01\tiny±0.94 & 89.38\tiny±0.18 & 0.64& 60.48\tiny±0.22 & 90.43\tiny±0.08 & 0.96\\
        High-Frequency Dominance (c)& 41.72\tiny±10.88 & 88.86\tiny±0.40 & 0.64& 57.94\tiny±1.55 & 90.08\tiny±0.17 & 0.96\\
        {{\bf Mutual Integration (d) (proposed \method)}}&{{\bf 44.02\tiny±3.28 }}& {{\bf89.51\tiny±0.35}} & {{\bf0.64}}&{{\bf 61.16\tiny±0.85 }}& {{\bf90.84\tiny±0.43}} & {{\bf0.96}}\\
        \midrule
        Mutual Integration + Width Reduction Only& 47.50\tiny±6.33 & 87.48\tiny±1.12 & 1.39& 58.06\tiny±0.87 & 87.97\tiny±0.29 & 2.09 \\
        Mutual Integration + Depth Reduction Only& 45.32\tiny±3.21 & 89.52\tiny±0.21 & 2.53& 61.46\tiny±0.37 & 90.85\tiny±0.04 & 3.80\\
        \midrule
\end{tabular}
\label{tab:Ablation}
\end{table*}
\begin{figure} [t!]
     \centering
     \includegraphics[width=0.7\linewidth]{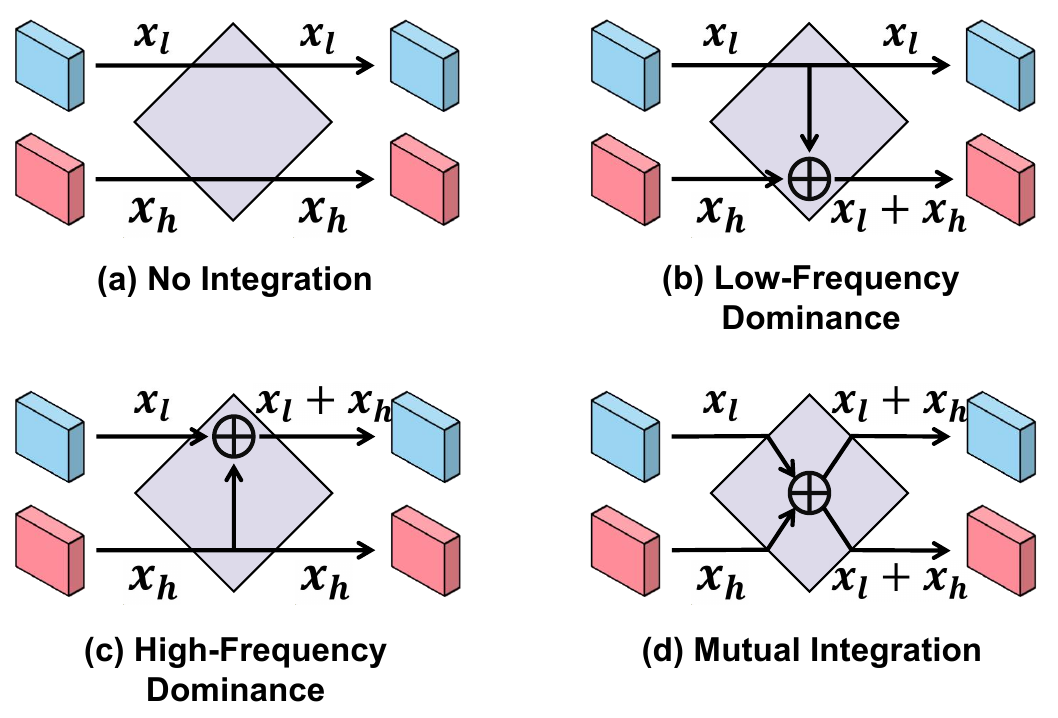}  
     \caption{Design choices for frequency integration operations in feature aggregators.}
\vspace{-0.1in}
\label{fig:Ablation}
\end{figure}

\subsection{Frequency Integration}
\label{sec:method:Integration}
While frequency decomposition replay aids the model in recalling both low-frequency and high-frequency image information, L-Net and H-Net are still susceptible to forgetting. To better facilitate the model's learning and recall of class-specific features, we need to integrate the image's global structure with its class-specific details. To address this, we design feature aggregators to integrate and exchange intermediate low-frequency and high-frequency features between L-Net and H-Net, enabling mutual enhancement and improved overall performance. In our design, feature integration takes place prior to each downsampling stage of the network, including the initial stage that does not involve downsampling. The number of feature aggregators corresponds to the number of stages in the lightweight network. We define the features extracted by L-Net and H-Net at the $i$-th stage as $x_l^i$ and $x_h^i$, respectively, which are provided to the feature aggregator. The aggregator then combines these features as follows:
\begin{equation}
\hat{x}_l^i = \hat{x}_h^i = x_l^i + x_h^i .
\end{equation}
The feature aggregator subsequently transmits the reprogrammed features, $\hat{x}_l^i$ and $\hat{x}_h^i$, to L-Net and H-Net, respectively, ensuring that each lightweight network effectively utilizes the distinct frequency information for improved feature extraction. We use addition to integrate frequency features~\cite{yang2022rep} and propose four design choices. Fig.~\ref{fig:Ablation} (a) presents the case where high-frequency and low-frequency features are not integrated. In this setting, L-Net and H-Net independently extract low-frequency and high-frequency features, which are concatenated only before the final classification layer. In Fig.~\ref{fig:Ablation} (b), for the low-frequency dominance design, we modify H-Net features using the integrated features (i.e., $x_l$ + $x_h$) while keeping L-Net features (i.e., $x_l$) unchanged. This highlights the importance of low-frequency features in CL. In Fig.~\ref{fig:Ablation} (c), for the high-frequency dominance design, we modify only L-Net features while keeping H-Net features unchanged, thereby assessing the significance of high-frequency features in CL. Fig.~\ref{fig:Ablation} (d) illustrates the mutual integration design, where both lightweight networks benefit from integrating the features. We conduct an ablation study, and provide the results in Tab.~\ref{tab:Ablation}. The results in rows 1 to 4 indicate that our mutual integration design outperforms all other cases, as the mutual integration of features between L-Net and H-Net enables a complementary enhancement of performance. Moreover, the comparison between rows 2 and 3 demonstrates that low-frequency dominance outperforms high-frequency dominance. This is likely due to the fact that low-frequency features capture the global structure of the image more effectively, thereby improving the learning ability of H-Net.

\section{Experiment}
\subsection{Experimental Setup}
\label{exp:setup}
\begin{table*}[tb]
\caption{Comparison with different CL methods. The optimal results are marked in bold, and shadowed lines indicate the results from our framework.}
\label{tab:all-datasets}
\centering
\renewcommand\arraystretch{1}	\setlength{\tabcolsep}{3.mm}
\scalebox{0.95}{
\begin{tabular}{ll|ccc|ccc}
\toprule
\multirow{2}{*}{{Buffer}} & \multirow{2}{*}{{Method}} & \multicolumn{3}{c|}{\textbf{Split CIFAR-10}} & \multicolumn{3}{c}{\textbf{Split Tiny-ImageNet}}  \\ \cmidrule{3-5} \cmidrule{6-8}
 &  &  {Class-IL($\uparrow$)} & {Task-IL($\uparrow$)} &  {Mem($\downarrow$)} & {Class-IL($\uparrow$)} & {Task-IL($\uparrow$)} & {Mem($\downarrow$)} \\ \midrule
 
\multirow{2}{*}{–} & JOINT & 92.20\tiny±0.15 & 98.31\tiny±0.12 & - & 59.99\tiny±0.19  & 82.04\tiny±0.10 & - \\
 & SGD & 19.62\tiny±0.05 & 61.02\tiny±3.33 & - & 7.92\tiny±0.26  & 18.31\tiny±0.68 & - \\  
\midrule
\multirow{3}{*}{–} & oEWC~\cite{schwarz2018progress} & 19.49\tiny±0.12 & 68.29\tiny±3.92 & 530MB & 7.58\tiny±0.10  & 19.20\tiny±0.31 & 970MB \\
 & SI~\cite{zenke2017continual} & 19.48\tiny±0.17 & 68.05\tiny±5.91 & 573MB & 6.58\tiny±0.31  & 36.32\tiny±0.13 & 1013MB \\
 & LwF~\cite{li2017learning} & 19.61\tiny±0.05 & 63.29\tiny±2.35 & 316MB & 8.46\tiny±0.22  & 15.85\tiny±0.58 & 736MB \\
\midrule
 & ER~\cite{robins1995catastrophic} & 29.42\tiny±3.53 & 86.36\tiny±1.43 & 497MB & 8.14\tiny±0.01 & 26.80\tiny±0.94 & 1333MB \\ 
 & DER++~\cite{buzzega2020dark} & 42.15\tiny±7.07 & 83.51\tiny±2.48 & 646MB & 8.00\tiny±1.16 & 23.53\tiny±2.67 & 1889MB \\ 
 & ER-ACE~\cite{caccia2021new} & 40.96\tiny±6.00 & 85.78\tiny±2.78 & 502MB & 6.68\tiny±2.75 & 35.93\tiny±2.66 & 1314MB \\ 
 \multirow{-4}{*}{50}& CLS-ER~\cite{arani2022learning} & 45.91\tiny±2.93 & 89.71\tiny±1.87 & 1016MB & 11.09\tiny±11.52 & 40.76\tiny±9.17 & 3142MB \\
 \midrule
 \rowcolor[gray]{.9} & \method-ER & 34.26\tiny±4.10 & 86.65\tiny±1.27 & 128MB & 7.58\tiny±0.01 & 35.65\tiny±0.67 & 278MB \\ 
 \rowcolor[gray]{.9} & \method-DER++ & 45.92\tiny±3.43 & 84.86\tiny±2.11 & 152MB & 8.55\tiny±0.19 & 31.12\tiny±0.35 & 374MB \\ 
 \rowcolor[gray]{.9} & \method-ER-ACE & 49.01\tiny±2.40 & 86.89\tiny±1.24 & 128MB & 9.02\tiny±0.81 & 39.34\tiny±0.83 & 278MB \\ 
  \rowcolor[gray]{.9}  \multirow{-4}{*}{50}& \method-CLS-ER &\textbf{53.40\tiny±5.77} & \textbf{89.91\tiny±0.25}& 309MB & \textbf{13.62\tiny±1.46} & \textbf{44.63\tiny±4.33} & 675MB \\
\midrule
 & ER~\cite{robins1995catastrophic} & 38.49\tiny±1.68 & 89.12\tiny±0.92 & 497MB & 8.30\tiny±0.01 & 34.82\tiny±6.82 & 1333MB \\ 
 & DER++~\cite{buzzega2020dark} & 53.09\tiny±3.43 & 88.34\tiny±1.05 & 646MB & 11.29\tiny±0.19 & 32.92\tiny±2.01 & 1889MB \\ 
 & ER-ACE~\cite{caccia2021new} & 56.12\tiny±2.12 & 90.49\tiny±0.58 & 502MB & 11.09\tiny±3.86 & 41.85\tiny±3.46 & 1314MB \\ 
 \multirow{-4}{*}{125}& CLS-ER~\cite{arani2022learning} & 53.57\tiny±2.73 & 90.75\tiny±2.76 & 1016MB & 16.35\tiny±4.61 & 46.11\tiny±7.69 & 3142MB \\
  \midrule
 \rowcolor[gray]{.9} & \method-ER & 44.02\tiny±3.28 & 89.51\tiny±0.35 & 128MB & 8.11\tiny±0.04 & 39.57\tiny±0.97 & 278MB \\ 
 \rowcolor[gray]{.9} & \method-DER++ & 55.13\tiny±1.64 & 88.10\tiny±0.67 & 152MB & 11.85\tiny±0.78 & 39.39\tiny±0.31 & 374MB \\ 
 \rowcolor[gray]{.9} & \method-ER-ACE & 56.49\tiny±1.70 & 89.61\tiny±0.29 & 128MB & 14.24\tiny±0.59 & 44.76\tiny±0.37 & 278MB \\ 
  \rowcolor[gray]{.9}  \multirow{-4}{*}{125}& \method-CLS-ER & \textbf{61.16\tiny±0.85} & \textbf{90.84\tiny±0.43} & 309MB & \textbf{17.30\tiny±1.42} & \textbf{52.54\tiny±0.80} & 675MB \\
\bottomrule
\end{tabular}
}
\end{table*}
\paragraph{Datasets.}
Following the benchmark setting~\cite{buzzega2020dark}, we use three datasets: CIFAR-10~\cite{krizhevsky2009learning}, Tiny ImageNet~\cite{deng2009imagenet}, and ImageNet-R~\cite{hendrycks2021many}. We split CIFAR-10 into 5 tasks, each consisting of 2 classes, and both Tiny ImageNet and ImageNet-R into 10 tasks, each consisting of 20 classes. Details of these datasets are provided in Appendix B.
\paragraph{Evaluation metrics.}
We evaluate the final model's performance using the average accuracy across all tasks in both the Task Incremental Learning (Task-IL) and Class Incremental Learning (Class-IL) settings. In the Task-IL setting, task identity is known during testing, allowing for per-task classification. In contrast, the Class-IL setting does not provide task identity during inference, requiring the model to classify across all learned classes, making it a more challenging scenario. The average accuracy is calculated as follows:
\begin{equation}
  ACC_t = \frac{1}{t}\sum_{\tau=1}^{t}A_{t,\tau},
  \label{eq:average accuracy}
\end{equation}
where $A_{t,\tau}$ represents the testing accuracy for task $\tau$ when the model has completed learning task $t$. Moreover, we evaluate the peak memory footprint, training time and training FLOPs~\cite{wang2022sparcl} to demonstrate the efficiency of each method. Additional experimental results, including measures of forgetting, stability, plasticity, and task-specific accuracy, can be found in Appendix E.
\paragraph{Baselines.}
We compare \method with several representative rehearsal-based methods, including ER~\cite{robins1995catastrophic}, DER++~\cite{buzzega2020dark}, ER-ACE~\cite{caccia2021new} and CLS-ER~\cite{arani2022learning}, as well as well-known regularization-based methods, such as oEWC~\cite{schwarz2018progress}, SI~\cite{zenke2017continual} and LwF~\cite{li2017learning}. Additionally, we incorporate two non-continual learning benchmarks: SGD (as the lower bound) and JOINT (as the upper bound).
\paragraph{Implementation Details.}
We use ResNet18 as the backbone network and train it from scratch on each dataset. All models are trained using the Stochastic Gradient Descent (SGD) optimizer with a batch size of 32. Each experiment is repeated 10 times with different initializations, and the results are averaged. More details are in Appendix D.
\subsection{Main Result}
\paragraph{Comparison with Effective CL Methods.}
Tab.~\ref{tab:all-datasets} presents the average accuracy for both Class-IL and Task-IL settings. As demonstrated in Tab.~\ref{tab:all-datasets}, \method utilizes frequency decomposition to compress both the input and network structure, resulting in a reduction of peak memory footprint by up to 80\% when combined with DER++. Additionally, by integrating frequency information and efficiently using storage resources, \method, in combination with CLS-ER, achieves an accuracy improvement of up to 7.49\% over SOTA methods. The frequency decomposition and integration employed by \method significantly improve the accuracy of all rehearsal-based methods while reducing peak memory footprint. Furthermore, \method demonstrates adaptability as a unified framework, highlighting its potential for integration with various CL methods, achieving impressive performance while reducing backbone parameters by 78\%.
\begin{figure} [t!]
     \centering
     \includegraphics[width=1\linewidth]{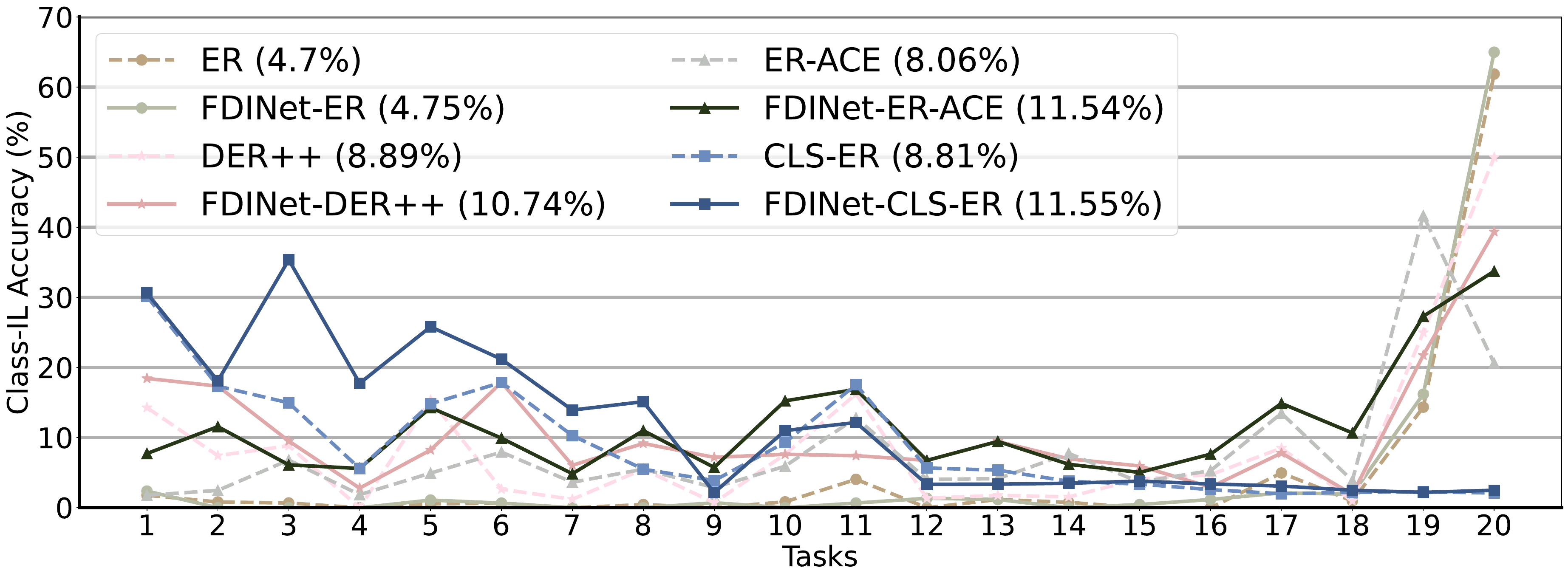}  
     \caption{Comparison of Class-IL accuracy of different methods on the Split ImageNet-R dataset. The values in parentheses in the legend indicate the average accuracy.}
\vspace{-0.1in}
\label{fig:imgr}
\end{figure}

To demonstrate task-specific accuracy across different comparison methods, we conduct experiments on the ImageNet-R dataset, evaluating the Class-IL accuracy for each task with a buffer size of 500. Fig.~\ref{fig:imgr} shows the accuracy of individual tasks at the end of CL training. The results illustrates that \method consistently improves the accuracy of various rehearsal-based methods, highlighting its robustness and general applicability. Furthermore, methods combined with \method generally achieve superior performance across almost all tasks, indicating the effectiveness of \method in mitigating forgetting and ensuring stable learning throughout the CL process.
\paragraph{Comparison with Efficient CL Methods.}
\begin{table}[t]
\centering
\renewcommand\arraystretch{1.2}\setlength{\tabcolsep}{1.1mm}
\caption{Comparison of \method and efficient CL methods on Split CIFAR-10 (Buffer Size: 50, SparCL Sparsity Ratio: 0.75). The optimal results are marked in bold.}
\begin{tabular}{cccc}
 		\hline
		\multirow{3}{*}{\small{{\bf  Method}}}&\multicolumn{3}{c}{\small{{\bf Combined with CLS-ER}}}\\
		\cline{2-4}
		&\multirow{2}{*}{\small{Class-IL($\uparrow$)}}&\multirow{2}{*}{\small{Task-IL($\uparrow$)}}& \small{FLOPs Train} \\
		&&&\small{$\times 10^{15}$($\downarrow$)}\\
        \hline
        CLS-ER~\cite{arani2022learning} & 45.91\tiny±2.93 & 89.71\tiny±1.87 & 16.67 \\
        \hline
        SparCL~\cite{wang2022sparcl} & 49.31\tiny±2.47 & 87.38\tiny±1.73 & 4.15\\
        CLFD~\cite{liu2024continual}& 50.13\tiny±3.67 & 85.30\tiny±1.01 & 4.17\\
        {{\bf \method}}&\textbf{53.40\tiny±5.77} & \textbf{89.91\tiny±0.25} & {{\bf0.96}}\\
        \hline
\end{tabular}
\label{tab:efficient}
\end{table}
Tab.~\ref{tab:efficient} shows the results of combining both our method and SOTA efficient CL methods with CLS-ER. The results demonstrates that all efficient CL methods improve the accuracy of CLS-ER. SparCL mitigates model overfitting primarily through sparse training, while CLFD and \method achieve performance gains by efficiently utilizing different frequency components of images and enhancing storage resource utilization. Regarding training acceleration, SparCL and CLFD accelerate the CL process by focusing solely on either the network structure or the input. In contrast, \method compresses both the network structure and the input, achieving the highest accuracy improvement with the lowest training FLOPs.

\begin{figure} [t!]
     \centering
     \includegraphics[width=0.8\linewidth]{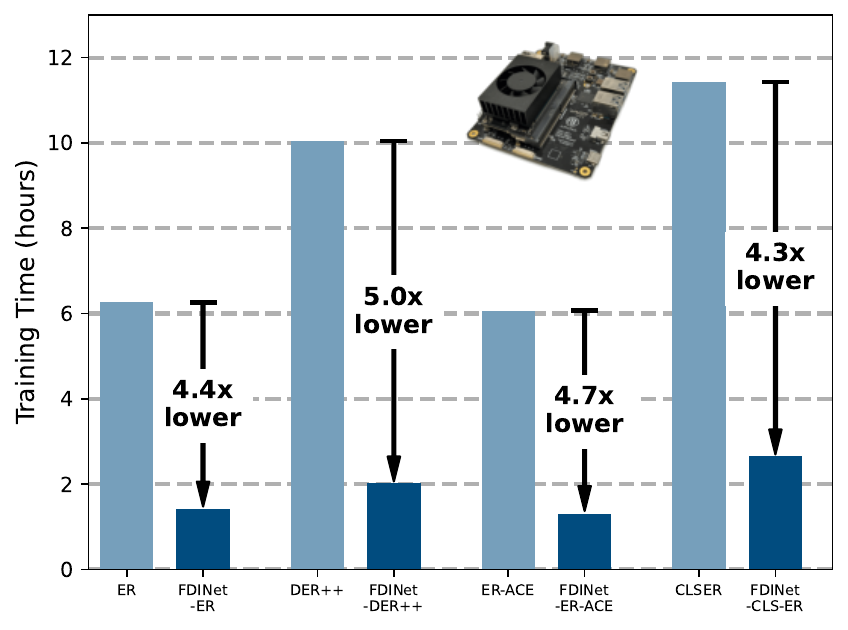}  
     \caption{Comparison of different methods for Split CIFAR-10 dataset using Nvidia Jetson Orin NX with a buffer size of 125.}
\vspace{-0.1in}
\label{fig:time}
\end{figure}
\subsection{Discussion}
\begin{figure*} [t!]
     \centering
     \includegraphics[width=0.83\linewidth]{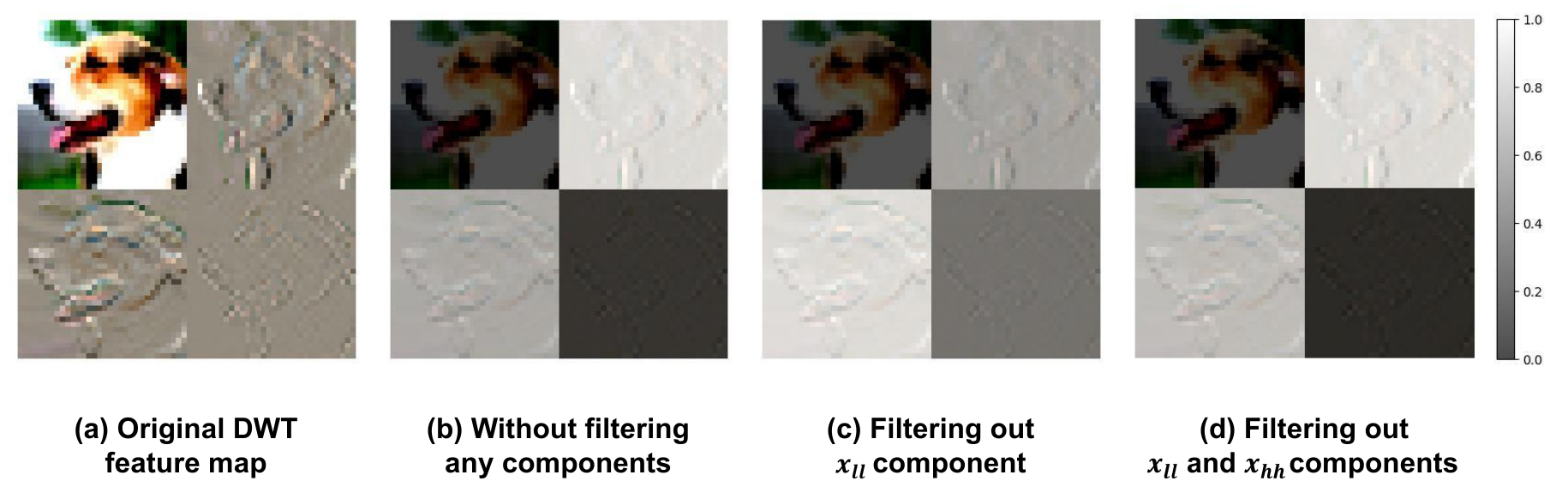}  
     \caption{The preference of different convolutions for various frequency components in DWT feature maps. Brighter regions indicate higher weights, while darker regions represent lower weights.}
\vspace{-0.1in}
\label{fig:filter}
\end{figure*}
\paragraph{Training Times.}
\label{sec:time}
 In this section, we validate its practical acceleration on edge devices using the NVIDIA Ampere GPU and Octa-core Arm CPU on the NVIDIA Jetson Orin NX 16GB platform. Experiments are performed on the Split CIFAR-10 dataset with a buffer size of 125. Fig.~\ref{fig:time} shows that our method significantly improves the training speed of rehearsal-based methods, achieving up to a 5$\times$ speedup when combined with DER++. Notably, our method accomplishes this acceleration using PyTorch alone, without additional optimization support\footnote{We analyze the theoretical training speedup of \method in Appendix A.}.
\begin{table}[t]
\centering
\renewcommand\arraystretch{1.2}\setlength{\tabcolsep}{1.7mm}
\caption{Comparison of different frequency components on the Split CIFAR-10 dataset with a buffer size of 50. The results of our framework are highlighted in bold.}
\begin{tabular}{ccc}
 \hline
 \multirow{2}{*}{{\bf  Method}}&\multicolumn{2}{c}{{\bf Combined with CLS-ER}}\\
 \cline{2-3}
 & Class-IL($\uparrow$) & Task-IL($\uparrow$)\\
 \hline
 $X_{ll}$ & 48.92\tiny±5.09 & 87.65\tiny±1.52 \\
 $X_{lh}$ & 52.06\tiny±8.31 & 88.72\tiny±2.41 \\
 $X_{hl}$ & 50.67\tiny±0.52 & 88.78\tiny±0.29  \\
 $X_{hh}$ & 49.31\tiny±2.54 & 88.53\tiny±0.20  \\
 \hline
 1 $\times$ 1 Conv & 52.47\tiny±5.30 & 88.80\tiny±0.54 \\
  1 $\times$ 1 Conv w/o $X_{ll},X_{hh}$ & 52.22\tiny±4.83 & 88.43\tiny±0.79 \\
  {\bf 1 $\times$ 1 Conv w/o $X_{ll}$} & {\bf 53.40\tiny±5.77} & {\bf89.91\tiny±0.25} \\
 \hline
\end{tabular}
\label{tab:dif}
\end{table}
\paragraph{Utilization of Different Frequency Components.}
\label{sec:filter}
Fig.~\ref{fig:high_low} demonstrates that both low-frequency and high-frequency components of images preserve class-discriminative features. However, a critical question arises: \textit{Can CL benefit from decoupled learning of low-frequency and high-frequency features?} To investigate this, we first maintain the input to L-Net unchanged, given the established benefits of low-frequency components for CL~\cite{liu2024continual}. Subsequently, we modify the input to H-Net by directly providing frequency components derived from the DWT ($X_{ll}$, $X_{lh}$, $X_{hl}$, $X_{hh}$) to test their individual contributions. The results, presented in rows 1 to 4 of Tab.~\ref{tab:dif}, show that using $X_{ll}$ and $X_{hh}$ does not yield satisfactory performance. This can be attributed to: (1) CL solely on low-frequency $X_{ll}$ leads to a loss of class-specific details, resulting in \textit{catastrophic forgetting}. (2) High-frequency $X_{hh}$ fails to accurately capture class-specific information, causing suboptimal CL performance. Conversely, $X_{lh}$ and $X_{hl}$ achieve competitive results, suggesting effective capture of class-discriminative textures. The frequency-decoupled rehearsal strategy enhances holistic representations by preserving low-frequency information while mitigating the forgetting of class-specific details through the retention of high-frequency information.

To further verify the impact of different frequency components on \method, we use a pointwise convolution to fuse all frequency components (1 × 1 Conv in Tab.~\ref{tab:dif}) and compare it to filtering $X_{ll}$ (1 × 1 Conv w/o $X_{ll}$ in Tab.~\ref{tab:dif}) and filtering both $X_{ll}$ and $X_{hh}$ (1 × 1 Conv w/o $X_{ll}, X_{hh}$ in Tab.~\ref{tab:dif}). To prevent forgetting, we freeze all pointwise convolutions after the completion of the first task. The results are shown in rows 5 to 7 of Tab.~\ref{tab:dif}, indicating that all three methods achieve competitive results.To explain the underlying mechanisms, we compute the normalized sum of the absolute values of the weights for each frequency component in the pointwise convolution (assigning a value of zero for filtered components) and visualize them superimposed on the original DWT feature maps, as shown in Fig.~\ref{fig:filter}. From the visualization, we observe that regardless of whether $X_{ll}$ or $X_{hh}$ is removed, the pointwise convolution consistently focuses on the $X_{lh}$ and $X_{hl}$ components during training. This observation aligns with our prior analysis and explains why all three methods produce similar results.

\paragraph{Impact of Different Buffer Sizes.}
\begin{figure} [t!]
     \centering
     \includegraphics[width=1\linewidth]{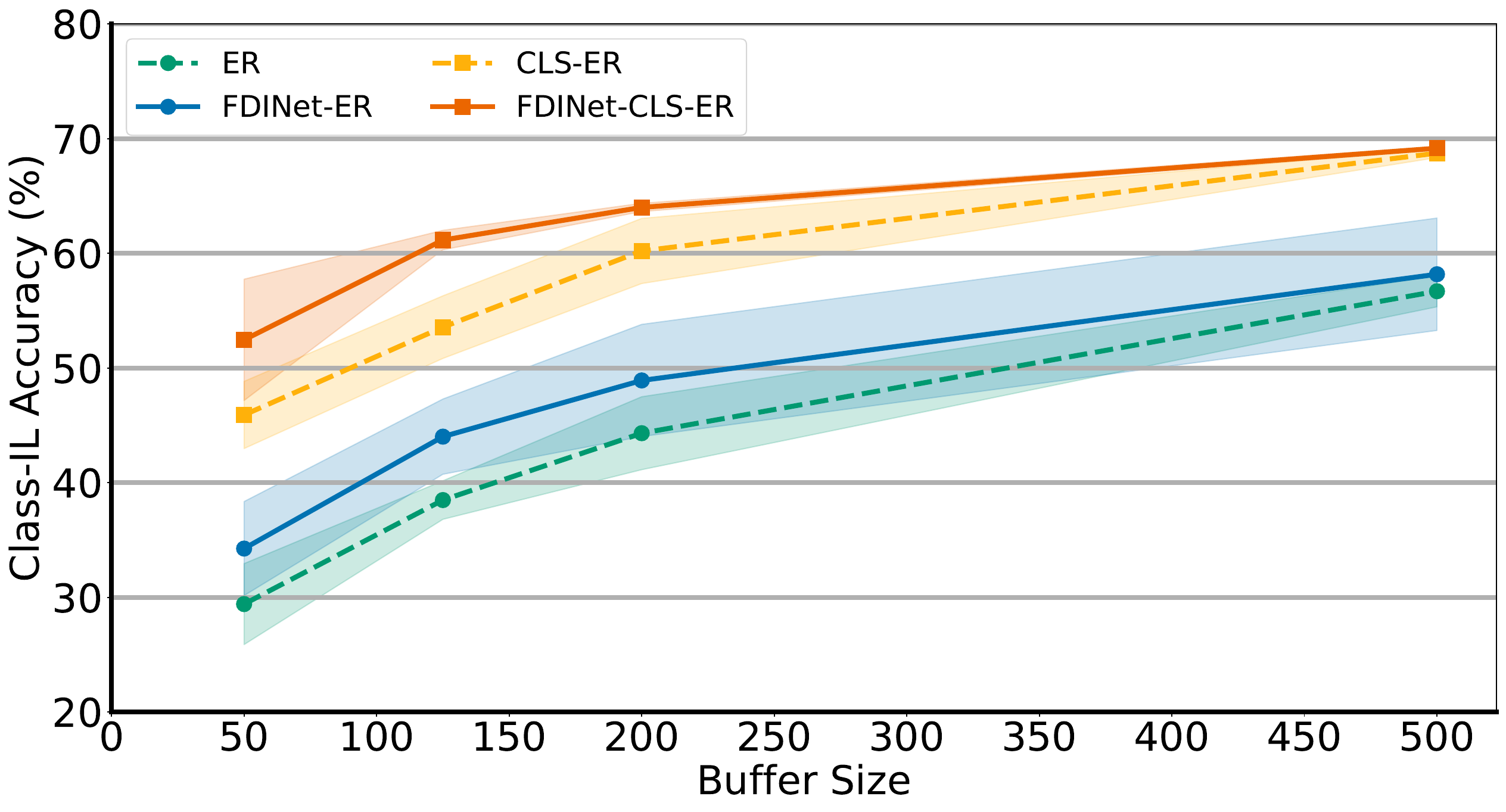}  
     \caption{Performance comparison with different buffer sizes on the Split CIFAR-10 dataset.}
\vspace{-0.1in}
\label{fig:buffer}
\end{figure}
Since rehearsal-based methods rely on replay buffers to mitigate forgetting, we investigate the relationship between \method and buffer size on the Split CIFAR-10 dataset, with results shown in Fig.~\ref{fig:buffer}. As observed, larger buffer sizes generally lead to better performance. Furthermore, when integrated with ER and CLS-ER, our method consistently improves accuracy across all buffer settings, highlighting the stability of \method.
\section{Conclusion}
We propose \method, a novel framework inspired by the human perceptual system, which synergistically optimizes both training efficiency and accuracy for rehearsal-based methods through frequency decomposition and integration. \method compresses both the network input and structure, thereby enhancing training efficiency, and proposes a frequency-decoupled replay strategy to effectively compensate for performance loss caused by compression, thus improving accuracy. Our experiments demonstrate that \method significantly enhances the accuracy of other rehearsal-based methods while reducing memory footprint and accelerating training.
{
    \small
    \bibliographystyle{ieeenat_fullname}
    \bibliography{main}
}

\end{document}